\pdfoutput=1
\RequirePackage{fix-cm}
\PassOptionsToPackage{table}{xcolor}
\documentclass[11pt,nologo]{alta}
\usepackage[round,authoryear]{natbib}

\usepackage{amsmath,amsfonts,bm}

\def\eqref#1{\cref{#1}}

\def\1{\bm{1}}

\DeclareMathAlphabet{\mathsfit}{\encodingdefault}{\sfdefault}{m}{sl}
\SetMathAlphabet{\mathsfit}{bold}{\encodingdefault}{\sfdefault}{bx}{n}

\usepackage{latexsym}
\usepackage{amsmath}
\usepackage{subcaption}
\usepackage{enumitem}
\usepackage{tabularx}
\usepackage{booktabs}
\usepackage{multirow}
\usepackage{graphicx}
\usepackage{makecell}
\usepackage{soul}
\usepackage[table]{xcolor}
\usepackage{fontawesome5} 
\tcbuselibrary{most}
\definecolor{promptPurple}{HTML}{5D1548} 
\usepackage{amsfonts}   
\usepackage{amsmath}               

\usepackage{amsthm}                
\usepackage{mathtools}             
\theoremstyle{plain}

\theoremstyle{definition}

\usepackage{xcolor}

\usepackage[table]{xcolor}
\definecolor{GainBase}{RGB}{70,170,70}
\definecolor{DropBase}{RGB}{220,80,80}
\newcommand{\gainN}[2]{%
  \colorbox{GainBase!#1!white}{\scriptsize +#2}%
}
\newcommand{\dropN}[2]{%
  \colorbox{DropBase!#1!white}{\scriptsize -#2}%
}
\newtcolorbox{promptbox}[2][]{%
    enhanced,
    breakable,     
    colback=white,      
    colframe=promptPurple, 
    coltitle=white,     
    fonttitle=\bfseries\large,
    title={\faInfoCircle\ \ #2}, 
    arc=3mm,            
    boxrule=1.5pt,      
    left=5pt, right=5pt, top=5pt, bottom=5pt,
    overlay first={
        \draw[promptPurple, line width=1.5pt] (frame.south west) -- (frame.south east); 
    },
    overlay middle={
        \draw[promptPurple, line width=1.5pt] (frame.north west) -- (frame.north east);
        \draw[promptPurple, line width=1.5pt] (frame.south west) -- (frame.south east);
    },
    overlay last={
        \draw[promptPurple, line width=1.5pt] (frame.north west) -- (frame.north east);
    },
    #1
}

\definecolor{figpurple}{HTML}{68579E}        
\colorlet{pivotPurpleDark}{figpurple}        
\definecolor{pivotPurpleLight}{HTML}{ECEAF6}  
\newtcolorbox{takeawaybox}{
  enhanced,
  colback=white,
  frame hidden,
  borderline north={0.7pt}{0pt}{figpurple},
  borderline south={0.7pt}{0pt}{figpurple},
  boxrule=0pt,
  arc=0pt,
  left=0pt,
  right=0pt,
  top=1.2mm,
  bottom=1.2mm,
  before upper={\textcolor{figpurple}{\textbf{Takeaway.}}\enspace},
  before skip=10pt,
  after skip=6pt
}

\definecolor{casePurpleDark}{HTML}{66589F}
\definecolor{casePurpleLight}{HTML}{F9F8FB}
\newtcolorbox{guidancecasebox}{
  colback=casePurpleLight,
  colframe=casePurpleDark,
  colbacktitle=casePurpleLight,
  coltitle=pivotPurpleDark,
  fontupper=\small,
  title={Why does middle intervention help?},
  fonttitle=\bfseries,
  boxrule=0.7pt,
  arc=1.2mm,
  left=1.8mm,
  right=1.8mm,
  top=1mm,
  bottom=1mm,
  before skip=10pt,
  after skip=8pt,
  before upper={\setlength{\parskip}{\smallskipamount}}
}

\definecolor{hl_yellow}{HTML}{FFF3A8}
\definecolor{hl_blue}{HTML}{DAE8FC}
\definecolor{hl_pink}{HTML}{FFE3ED}
\definecolor{hl_green}{HTML}{D4F5E3}

\newcommand{\hly}[1]{%
  {\sethlcolor{hl_yellow}\hl{#1}}%
}

\newcommand{\hlp}[1]{%
  {\sethlcolor{hl_pink}\hl{#1}}%
}
\newcommand{\hlg}[1]{%
  {\sethlcolor{hl_green}\hl{#1}}%
}

\usepackage[T1]{fontenc}

\usepackage[utf8]{inputenc}

\usepackage{microtype}

\usepackage{inconsolata}

\usepackage{graphicx}
\usepackage{wrapfig}   
\usepackage{xspace}
\usepackage{algorithm}
\usepackage{algpseudocode}
\usepackage{pifont}

\usepackage{tabularray}
\UseTblrLibrary{booktabs}

\newif\ifreviewnotes
\reviewnotestrue
\usepackage[textsize=scriptsize]{todonotes}

\newcommand{\method}{\textsc{PIVOT}\xspace}

\definecolor{studentblue}{HTML}{356AA0}
\definecolor{teacherpurple}{HTML}{6C5CE7}
\definecolor{rewardgreen}{HTML}{2A9D8F}
\definecolor{softgray}{HTML}{F3F5F7}
\definecolor{warmorange}{HTML}{E07A5F}
\definecolor{figteal}{HTML}{00806D}


\usepackage{hyperref}
\usepackage{url}

\usepackage[nameinlink]{cleveref}
\crefname{section}{Section}{Sections}
\crefname{table}{Tab.}{Tabs.}
\crefname{figure}{Fig.}{Figs.}
\crefname{theorem}{Theorem}{Theorems}
\crefname{lemma}{Lemma}{Lemmas}
\crefname{algorithm}{Alg.}{Algs.}
\crefname{appendix}{App.}{Apps.}
\crefname{equation}{Eq.}{Eqs.}

\usepackage{pivot_arxiv}

\title{PIVOT: Progressive Intervention\\via Online Teacher Correction}
\author{Zhihao Yang\textsuperscript{1,4,*}\and
  Boyuan Zheng\textsuperscript{2,*}\and
  Puhui Tan\textsuperscript{2,*}\and
  Yukun Chen\textsuperscript{1}\and
  Kuan Zhang\textsuperscript{2}\\[2pt]
  Longze Chen\textsuperscript{1}\and
  Jiaming Li\textsuperscript{1}\and
  Renke Shan\textsuperscript{1}\and
  Lu Wang\textsuperscript{2,\dag}\and
  Xianpei Han\textsuperscript{3,\dag}}
\affiliation{\textsuperscript{1}University of Chinese Academy of Sciences\quad
  \textsuperscript{2}Tsinghua University\\
  \textsuperscript{3}Chinese Information Processing Laboratory, Institute of Software, Chinese Academy of Sciences\\
  \textsuperscript{4}School of Advanced Interdisciplinary Sciences, University of Chinese Academy of Sciences\\[3pt]
  \textsuperscript{*}Equal contribution\quad
  \textsuperscript{\dag}Corresponding authors\\
  \href{mailto:yangzhihao25@mails.ucas.ac.cn}{\texttt{yangzhihao25@mails.ucas.ac.cn}}\quad
  \href{mailto:zhengby25@mails.tsinghua.edu.cn}{\texttt{zhengby25@mails.tsinghua.edu.cn}}}
\preprintlabel{}
\headeraffiliation{}
\runningtitle{PIVOT: Progressive Intervention via Online Teacher Correction}
\hypersetup{pdftitle={PIVOT: Progressive Intervention via Online Teacher Correction},
 pdfauthor={Zhihao Yang, Boyuan Zheng, Puhui Tan, Yukun Chen, Kuan Zhang, Longze Chen, Jiaming Li, Renke Shan, Lu Wang, Xianpei Han},
 pdfsubject={Teacher intervention and token-level supervision for RLVR},
 pdfkeywords={PIVOT, RLVR, teacher intervention, reasoning}}

\begin{document}
\begin{abstract}
Reinforcement learning from verifiable rewards (RLVR) is limited by sparse outcome-level feedback on hard reasoning tasks: when all rollouts in a group fail, rewards provide no learning signal.
Stronger teachers can provide additional guidance, but this raises a central question: how can such signals be incorporated into RLVR to support successful rollouts while remaining learnable for the student?
We distinguish two complementary mechanisms: \emph{trajectory-level intervention}, which improves access to reward-bearing trajectories, and \emph{token-level supervision}, which provides dense targets to guide student learning.
Our controlled experiments show that either mechanism alone can remain insufficient, motivating their combination.
We therefore introduce \textbf{\method}, which inserts teacher-generated corrections into student rollouts and applies token-level KL supervision to these segments alongside the RLVR objective.
Beyond their combination, we systematically analyze where to intervene and how to supervise, yielding a recipe for effective guidance.
Across hard logic games and mathematical reasoning benchmarks, \method consistently outperforms existing methods, nearly doubling mean avg@8 on logic games from 19.5\% for the best-performing baseline to 35.6\%.

\end{abstract}
\maketitle
\suppressfloats[t]
\section{Introduction}

\begin{figure*}[t]
\centering
\includegraphics[width=0.95\textwidth]{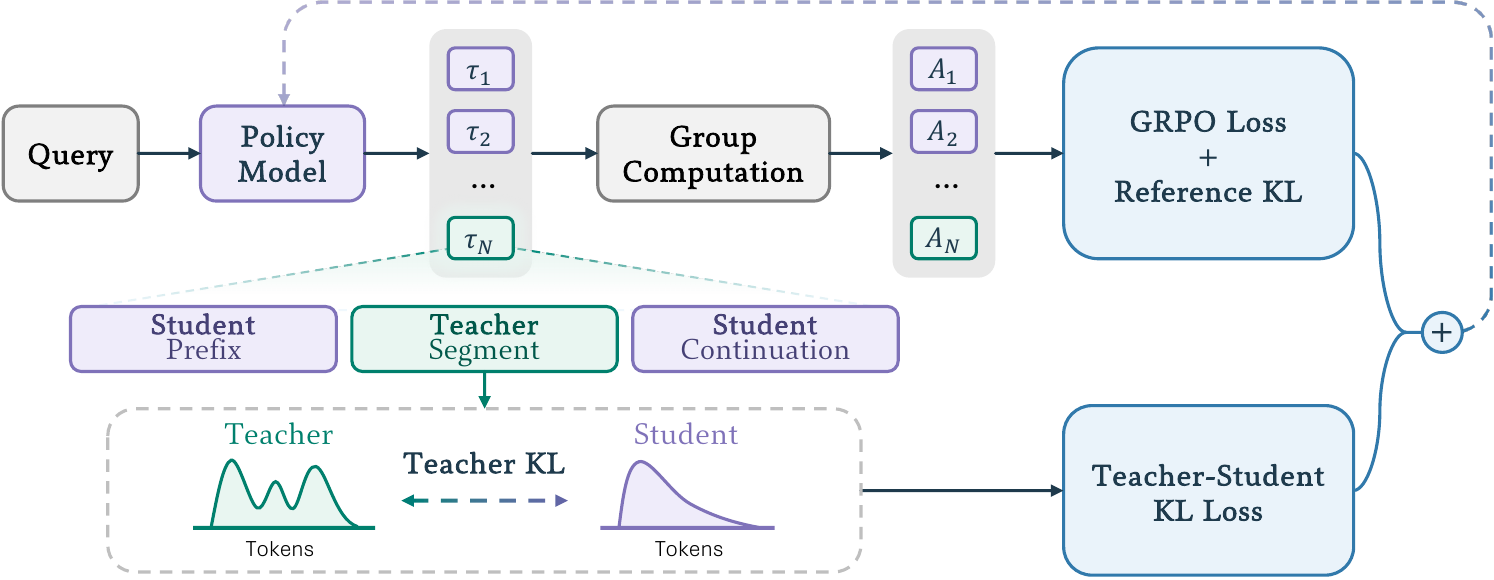}
\caption{
Overview of \method{}.
For each query, the policy model samples a group of rollouts, including teacher-assisted trajectories.
In a teacher-assisted rollout, the student first generates a prefix, the teacher provides a correction segment, and the student then continues generation.
The completed trajectory is optimized by the verifier-based GRPO objective, while the teacher-generated segment provides dense token-level supervision through a teacher-student KL loss.
}
\label{fig:pivot_framework}
\end{figure*}

Reinforcement learning from verifiable rewards (RLVR) has become a dominant paradigm for eliciting reasoning in large language models \citep{shao2024deepseekmath, lambert2024tulu, guo2025deepseek}.
Despite its success, recent studies suggest that RLVR primarily amplifies reasoning behaviors present in the base model, raising questions about its ability to acquire new capabilities \citep{yue2025does, zhao2025echo}.
This limitation is particularly pronounced on hard problems for which the initial policy rarely generates a correct solution \citep{yan2026learning, chen2025nudging, qu2026pope}.
In this regime, sparse outcome-level rewards provide little information for policy improvement: groups of unsuccessful rollouts receive identical zero rewards, leaving the policy without a reward-based learning signal \citep{wu2026learn, tao2025hybrid}.
These challenges have motivated the use of stronger teachers to provide guidance beyond verifier feedback.
However, introducing teacher signals does not by itself ensure that the student can learn from them.
This raises a central question: \emph{how can teacher guidance support \textbf{successful} rollouts while remaining \textbf{learnable} for the student?}

Existing approaches incorporate teacher signals into RLVR through two broad mechanisms.
\emph{Trajectory-level intervention} introduces teacher-generated trajectories or segments into training, improving access to reward-bearing trajectories \citep{guo2025deepseek, wen2025light, yan2025learning, qu2026pope}.
However, teacher-generated trajectories may differ substantially from the student's own rollout distribution, making the teacher's behavior difficult to learn even when these trajectories are successful.
\emph{Token-level supervision}, in contrast, provides dense supervision along student-generated rollouts using the teacher's predictive distributions \citep{xu2025kdrl}.
Yet when these rollouts deviate from a valid reasoning path, local corrective signals may not support a coherent recovery.
These limitations suggest that effective teacher guidance must shape both the trajectories used for training and the supervision derived from them.




We therefore introduce \textbf{P}rogressive \textbf{I}nter\textbf{v}ention via \textbf{O}nline \textbf{T}eacher Correction (PIVOT), a teacher-assisted RLVR framework that integrates teacher corrections into both rollout generation and policy optimization.
\method augments the rollout group with teacher-assisted trajectories, in which the teacher provides a correction based on a student-generated prefix and the student continues from the corrected context.
As illustrated in \cref{fig:pivot_framework}, verifier-based RLVR optimizes the resulting trajectories, while a teacher--student KL objective provides dense token-level supervision on the correction segments.
Our controlled experiments show that this joint use of teacher signals yields stronger learning than either intervention or token-level supervision alone.
We further examine how intervention placement and the KL objective affect the effectiveness of teacher guidance, informing a concrete recipe for where to intervene, how to supervise, and how much teacher guidance to apply.

We evaluate \method on hard logic games from SynLogic \citep{liu2026synlogic} and mathematical reasoning benchmarks.
On hard logic games, where RL-only training makes limited progress, \method nearly doubles mean avg@8 from 19.5\% for the best-performing baseline to 35.6\%.
On mathematical reasoning tasks, \method consistently outperforms the evaluated baselines across both in-distribution and out-of-distribution benchmarks.




Our contributions are threefold:
\begin{itemize}[leftmargin=1em, labelsep=0.4em]
    \item We identify two complementary mechanisms for teacher guidance in RLVR---\emph{trajectory-level intervention} and \emph{token-level supervision}---and show that either mechanism alone can remain insufficient for effective learning on hard reasoning tasks.

    \item We introduce \method, a teacher-assisted RLVR framework that integrates teacher corrections into both rollout generation and policy optimization, and provide a systematic analysis of how teacher guidance should be incorporated into training.

    \item We conduct extensive experiments and show that \method nearly doubles mean avg@8 over the best-performing baseline on hard logic games while consistently outperforming baselines across in-distribution and out-of-distribution mathematical reasoning benchmarks.
\end{itemize}

\section{Preliminaries}
\label{sec:preliminaries}

\paragraph{Problem setup.}
We consider reinforcement learning with verifiable rewards (RLVR) for a student policy $\pi_\theta$, with access to a teacher policy $\pi_T$ during training.
Given a prompt $\mathbf q$ drawn from the prompt distribution $\mathcal D$, a response $\tau=(\tau_1,\ldots,\tau_{|\tau|})$ receives a binary verifier reward $R(\mathbf q,\tau)\in\{0,1\}$, with $1$ indicating a correct response.
The student factorizes its response distribution as $\pi_\theta(\tau\mid\mathbf q)=\prod_{t=1}^{|\tau|}\pi_\theta(\tau_t\mid h_t)$, where $h_t=(\mathbf q,\tau_{<t})$ denotes the token context.
The objective is to maximize $\mathbb E_{\mathbf q\sim\mathcal D,\,\tau\sim\pi_\theta(\cdot\mid\mathbf q)}[R(\mathbf q,\tau)]$.

\paragraph{RLVR objective.}
Under Group Relative Policy Optimization (GRPO) \citep{shao2024deepseekmath}, groups of rollouts are sampled from $\pi_{\theta_{\mathrm{old}}}$ for each prompt, and their verifier rewards are normalized within each group to obtain advantages $A_i$.
For a batch of $N$ rollouts, the policy loss is averaged over all tokens:
\[
\mathcal L_{\mathrm{GRPO}}(\theta)
=-\frac{1}{\sum_{i=1}^{N}|\tau_i|}\sum_{i=1}^{N}
\sum_{t=1}^{|\tau_i|}\min\!\Bigl(
r_{i,t}(\theta)A_i,\,
\operatorname{clip}(r_{i,t}(\theta),1-\varepsilon,1+\varepsilon)A_i\Bigr),
\]
Here, $|\tau_i|$ is the rollout length, $\varepsilon>0$ is the clipping threshold, and $r_{i,t}(\theta)=\frac{\pi_\theta(\tau_{i,t}\mid h_{i,t})}{\pi_{\theta_{\mathrm{old}}}(\tau_{i,t}\mid h_{i,t})}$ is the importance ratio between the current and rollout policies.
The RLVR objective additionally includes a KL penalty to a fixed reference policy $\pi_{\mathrm{ref}}$:
\[
\mathcal L_{\mathrm{ref}}(\theta)
=\frac{1}{\sum_{i=1}^{N}|\tau_i|}\sum_{i=1}^{N}
\sum_{t=1}^{|\tau_i|}
\mathbb D_{\mathrm{KL}}\!\left(
\pi_\theta(\cdot\mid h_{i,t})\,\|\,\pi_{\mathrm{ref}}(\cdot\mid h_{i,t})\right).
\]
The RLVR loss is $\mathcal L_{\mathrm{GRPO}}+\beta_{\mathrm{ref}}\mathcal L_{\mathrm{ref}}$, with regularization weight $\beta_{\mathrm{ref}}\geq0$.

\paragraph{Token-level teacher supervision.}
Beyond outcome-level rewards, some methods use teacher predictions to provide dense token-level supervision~\citep{xu2025kdrl}.
This is commonly formulated as a forward- or reverse-KL loss between the teacher and student distributions:
\begin{align*}
\mathcal L_{\mathrm{KL}}^{\mathrm{forward}}(\theta;\mathcal S)
&=\frac{1}{|\mathcal S|}\sum_{(i,t)\in\mathcal S}
\mathbb D_{\mathrm{KL}}\!\left(\pi_T(\cdot\mid h_{i,t})\,\|\,\pi_\theta(\cdot\mid h_{i,t})\right),\\
\mathcal L_{\mathrm{KL}}^{\mathrm{reverse}}(\theta;\mathcal S)
&=\frac{1}{|\mathcal S|}\sum_{(i,t)\in\mathcal S}
\mathbb D_{\mathrm{KL}}\!\left(\pi_\theta(\cdot\mid h_{i,t})\,\|\,\pi_T(\cdot\mid h_{i,t})\right),
\end{align*}
where $\mathcal S$ denotes the token positions $(i,t)$ that receive teacher supervision.

\section{RLVR through Teacher Intervention and Supervision}

Teacher guidance can shape both the trajectories a student explores (i.e., \emph{trajectory-level intervention}) and the learning signals it receives (i.e., \emph{token-level supervision}).
Existing methods typically emphasize one of these two mechanisms.
Below, we examine why either mechanism alone can be insufficient for effective student learning and how combining them motivates the design of \method.

\subsection{Reward-Bearing Trajectories Are Not Necessarily Learnable}
\label{sec:reward_bearing_trajectories}

While the student may struggle to discover successful trajectories, \emph{trajectory-level intervention} by a stronger teacher provides access to reward-bearing reasoning paths.

Supervised fine-tuning (SFT) on these trajectories provides a warm start for subsequent RL \citep{guo2025deepseek,wen2025light}, while LUFFY incorporates successful teacher trajectories directly into RL training \citep{yan2025learning}.
POPE instead provides a teacher-generated prefix and lets the student generate the remaining continuation \citep{qu2026pope}.
These approaches differ in when and how teacher trajectories enter training, but all use teacher-generated trajectories or segments to improve access to successful reasoning paths.

\Needspace{23\baselineskip}
\begin{wrapfigure}{r}{0.42\textwidth}
\vspace{-8pt}
\centering
\includegraphics[width=\linewidth]{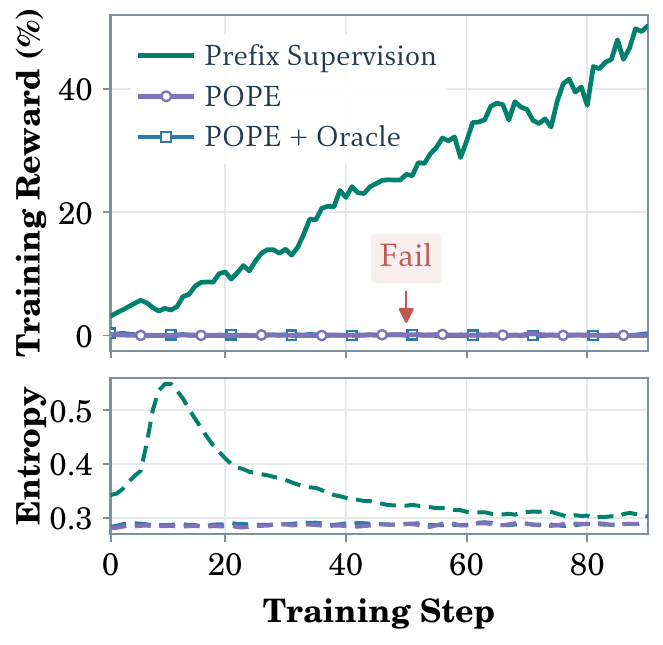}
\caption{
Prefix supervision vs. prefix-only guidance on Survo.
Supervised prefixes improve rewards, while POPE-style prefixes fail to yield comparable gains.
}
\label{fig:pope_survo_analysis}
\end{wrapfigure}
The success of a teacher trajectory, however, does not ensure that its behavior is readily learnable by the student.
Full teacher trajectories may differ substantially from the student's own rollout distribution, making them difficult imitation targets even when they receive positive rewards.
Providing only a teacher prefix allows the student to generate its own continuation, but learning to continue from a supplied prefix does not necessarily teach the student to produce the teacher decisions encoded in that prefix during independent generation.

\cref{fig:pope_survo_analysis} illustrates this distinction on Survo.
POPE yields limited gains even with stronger teacher prefixes.
We additionally consider oracle prefixes extracted from correct solutions to test whether prefix errors account for the limited gains.
Even with these prefixes, POPE makes little progress, whereas adding explicit teacher supervision on the provided prefixes improves training rewards.
This contrast suggests that prefix quality alone does not determine the usefulness of teacher guidance: the training objective must also enable the student to learn from the teacher-provided behavior.

\subsection{Dense Supervision Does Not Guarantee Trajectory Recovery}
\label{sec:dense_supervision}

Unlike trajectory-level intervention, \emph{token-level supervision} introduces teacher guidance directly into the learning objective.
KDRL~\citep{xu2025kdrl}, for example, combines GRPO with reverse-KL distillation along student-generated rollouts.
The teacher provides dense feedback at each token, but the student still determines the reasoning path and hence the contexts on which supervision is applied.

This dependence on student-generated contexts can complicate recovery from reasoning errors.
As observed in Trajectory-Refined Distillation (TRD)~\citep{jiang2026trajectory}, when a student prefix contains errors that require backtracking, the teacher may assign high probability both to continuing the student's reasoning and to initiating a correction.
Under such prefix failure, the resulting bimodal teacher mixture provides conflicting targets for token-level supervision.

\begin{figure*}[!htbp]
\centering
\includegraphics[width=\linewidth]{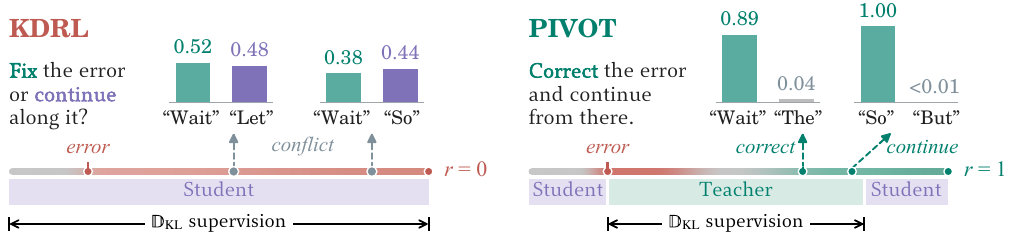}
\caption{
Dense supervision under KDRL and \method on a Survo rollout; insets show the teacher's top-2 next-token probabilities at supervised positions.
\textbf{Left:} on the student's failed trajectory, the teacher is split between fixing the error ({\color{figteal}``Wait''}) and continuing along it ({\color{figpurple}``Let''}, {\color{figpurple}``So''}), giving conflicting targets.
\textbf{Right:} the teacher corrects the error itself and continues from there, so its targets are unimodal whether it backtracks ({\color{figteal}``Wait''}) or proceeds ({\color{figteal}``So''}), and the trajectory reaches the reward.
}
\label{fig:bimodal_schematic}
\end{figure*}

\cref{fig:bimodal_schematic} (left) illustrates this issue on a Survo rollout trained with KDRL.
After an early error, the student continues along the wrong path and the rollout ends with zero reward.
At the supervised positions, teacher predictions repeatedly split between corrective tokens ({\color{figteal}``Wait''}) and tokens that continue the reasoning ({\color{figpurple}``Let''}, {\color{figpurple}``So''}), pulling the student in two directions at once.
Even when the teacher favors a correction, the remainder of the training sequence is still the student's erroneous continuation.
The student thus receives local corrective signals without observing how the proposed recovery unfolds.
When successful student rollouts are rare, training provides few examples of how to recover from an initial error and carry the revised reasoning through to a correct solution, limiting the ability of supervision alone to guide the student toward successful completion.

\subsection{PIVOT: Integrating Trajectory Guidance and Token-Level Supervision}

The preceding results suggest that the value of teacher guidance depends on both the reasoning paths it introduces and the supervision applied along those paths.
For methods relying solely on \emph{trajectory-level intervention}, \cref{fig:pope_survo_analysis} shows that adding a teacher-matching loss enables the student to learn more effectively from the provided prefixes.
For methods that apply \emph{token-level supervision} solely along student-generated rollouts, \cref{fig:bimodal_schematic} (right) shows how teacher intervention changes the supervision: the teacher generates a recovery path that becomes part of the training sequence and leads to a successful solution, providing largely unimodal targets both when initiating a correction ({\color{figteal}``Wait''}) and when continuing the revised reasoning ({\color{figteal}``So''}).

These observations motivate \method's joint use of teacher-generated segments for trajectory-level intervention and token-level supervision, allowing teacher guidance to shape both exploration and the learning objective.
We next describe the rollout construction and training objective.

\paragraph{Teacher-assisted rollout.}
For each prompt $\mathbf q$, we sample a group of rollouts $\mathcal G=\mathcal G_S\cup\mathcal G_T$, comprising student-only rollouts $\mathcal G_S$ and teacher-assisted rollouts $\mathcal G_T$, with sizes $G_S$ and $G_T$, respectively.
Student-only rollouts are sampled directly from $\pi_{\theta_{\mathrm{old}}}(\cdot\mid\mathbf q)$.
For teacher-assisted rollouts, the student first generates a reasoning prefix, the teacher continues from that prefix for a limited segment, and the student then resumes generation from the resulting context.
Writing the resulting rollout as $\tau=(u,v,w)$, the student generates $u$ and $w$, while the teacher generates $v$.
For an intervention position $\ell_u=|u|$ and teacher segment length $\ell_v=|v|$, this gives the sampling distribution
\[
p_{\theta_{\mathrm{old}},T}^{(\ell_u,\ell_v)}(\tau\mid\mathbf q)
=\pi_{\theta_{\mathrm{old}}}(u\mid\mathbf q)\,
\pi_T(v\mid\mathbf q,u)\,
\pi_{\theta_{\mathrm{old}}}(w\mid\mathbf q,u,v).
\]
Setting $\ell_u=0$ places the teacher segment at the beginning of the rollout; otherwise, the teacher conditions its intervention on a student-generated prefix.

\paragraph{Training objective.}
We combine the RLVR objective with a teacher-supervision loss applied only to teacher-generated tokens.
Let $\mathcal S_T$ denote the set of token positions belonging to the teacher-generated segments $v$ in the training batch.
The overall objective is
\[
\mathcal L(\theta)
=\mathcal L_{\mathrm{GRPO}}(\theta)
+\beta_{\mathrm{ref}}\mathcal L_{\mathrm{ref}}(\theta)
+\beta_{\mathrm T}\mathcal L_{\mathrm{KL}}(\theta;\mathcal S_T),
\]
where $\mathcal L_{\mathrm{KL}}$ is either the forward- or reverse-KL loss defined in \cref{sec:preliminaries}.
Together, the rollout construction and training objective define how \method uses teacher guidance, but leave open where to intervene, how to supervise, and how much guidance to provide, which we study next.

\section{Towards Effective Teacher Guidance: A Roadmap}\label{sec:effective_guidance}

We investigate where to intervene (\cref{sec:learnability}), how to formulate teacher supervision (\cref{sec:migration}), and how to configure the remaining guidance settings (\cref{sec:teacher_signal_calibration}) within \method. These analyses yield a practical training recipe, which we evaluate against existing methods in \cref{sec:main_results}.

\begin{table}[!b]
\definecolor{GainBase}{HTML}{00806D}
\colorlet{GainBase}{GainBase!65!white}
\definecolor{DropBase}{HTML}{BE5A51}
\setlength{\fboxsep}{1pt}
\renewcommand{\gainN}[2]{\colorbox{GainBase!#1!white}{\scriptsize\makebox[2.9em][r]{\strut +#2}}}
\renewcommand{\dropN}[2]{\colorbox{DropBase!#1!white}{\scriptsize\makebox[2.9em][r]{\strut -#2}}}
\newcommand{\placementvalue}[2]{\mbox{\makebox[2.6em][r]{#1}\hspace{0.15em}#2}}
\centering
\footnotesize
\setlength{\tabcolsep}{2pt}
\renewcommand{\arraystretch}{1.05}
\begin{tabularx}{\textwidth}{l*{6}{>{\centering\arraybackslash}X}}
\toprule
\multirow{2}{*}{Intervention}
& \multicolumn{2}{c}{Survo}
& \multicolumn{2}{c}{Minesweeper}
& \multicolumn{2}{c}{Kakurasu} \\
\cmidrule(lr){2-3} \cmidrule(lr){4-5} \cmidrule(lr){6-7}
& avg@8 & pass@8 & avg@8 & pass@8 & avg@8 & pass@8 \\
\midrule
\textcolor{figpurple}{Prefix} (\(\ell_u=0\))
& \placementvalue{45.90}{\phantom{\gainN{0}{0.00}}} & \placementvalue{87.00}{\phantom{\gainN{0}{0.00}}}
& \placementvalue{14.81}{\phantom{\gainN{0}{0.00}}} & \placementvalue{42.50}{\phantom{\gainN{0}{0.00}}}
& \placementvalue{0.57}{\phantom{\gainN{0}{0.00}}} & \placementvalue{3.20}{\phantom{\gainN{0}{0.00}}} \\
\midrule
\textcolor{figteal}{Middle} (\(\ell_u=512\))
& \placementvalue{50.70}{\gainN{30}{4.80}} & \placementvalue{\textbf{90.40}}{\gainN{25}{3.40}}
& \placementvalue{\textbf{19.25}}{\gainN{30}{4.44}} & \placementvalue{\textbf{55.50}}{\gainN{55}{13.00}}
& \placementvalue{3.52}{\gainN{25}{2.95}} & \placementvalue{\textbf{16.60}}{\gainN{55}{13.40}} \\
\textcolor{figteal}{Middle} (\(\ell_u=1024\))
& \placementvalue{49.88}{\gainN{28}{3.98}} & \placementvalue{86.40}{\dropN{10}{0.60}}
& \placementvalue{17.94}{\gainN{25}{3.13}} & \placementvalue{55.00}{\gainN{52}{12.50}}
& \placementvalue{\textbf{3.77}}{\gainN{26}{3.20}} & \placementvalue{15.20}{\gainN{50}{12.00}} \\
\textcolor{figteal}{Middle} (\(\ell_u=2048\))
& \placementvalue{\textbf{51.55}}{\gainN{33}{5.65}} & \placementvalue{87.20}{\gainN{10}{0.20}}
& \placementvalue{12.50}{\dropN{20}{2.31}} & \placementvalue{43.50}{\gainN{12}{1.00}}
& \placementvalue{3.32}{\gainN{24}{2.75}} & \placementvalue{14.00}{\gainN{47}{10.80}} \\
\bottomrule
\end{tabularx}
\caption{Effect of the intervention position \(\ell_u\) on hard logic games. \(\ell_u=0\) is prefix intervention.}
\label{tab:placement}
\end{table}

\subsection{Experimental Setup}
\label{sec:setup}
\paragraph{Tasks.}
We conduct experiments on hard logic games and mathematical reasoning.
For logic games, we train and evaluate on the hard subsets of Survo, Minesweeper, and Kakurasu from SynLogic~\citep{liu2026synlogic}, on which the base student almost never succeeds.
For mathematical reasoning, we pool problems from Polaris-53K~\citep{an2025polaris} and the math subset of Skywork-OR1-RL-Data~\citep{he2025skywork}.
We screen this pool with the base student Qwen3-4B, retaining challenging problems with low empirical success rates.
After removing problems with invalid answers and removing overlap with out-of-distribution benchmarks, we sample 30k training, 1k validation, and 2k held-out in-distribution test problems (Math-ID) from the filtered pool.
We also evaluate on the out-of-distribution benchmarks AIME 2024/2025~\citep{li2024numinamath} and MATH500~\citep{hendrycks2021measuring}.

\paragraph{Models and training.}
We use Qwen3-1.7B as the student for logic games and Qwen3-4B for mathematical reasoning, with Qwen3-4B-Thinking as the teacher in both settings~\citep{yang2025qwen3}.
All experiments use verl~\citep{sheng2025hybridflow} with GRPO~\citep{shao2024deepseekmath}.
By default, we sample six student-only and two teacher-assisted rollouts per prompt, all with thinking mode enabled.
The maximum response length is 8,192 tokens for logic games and 16,384 tokens for mathematical reasoning; by default, the teacher generates at most \(\ell_v=4{,}096\) tokens per intervention for logic games and \(8{,}192\) for math.

\paragraph{Evaluation.}
We evaluate all models with thinking mode enabled, reporting avg@8/pass@8 for the logic games, Math-ID, and MATH500, and avg@64/pass@64 for AIME 2024/2025.
Detailed descriptions of the baselines, datasets, and training hyperparameters are provided in \cref{appx:experimental_details}.

\subsection{Where Should the Teacher Intervene?}
\label{sec:learnability}
To study how intervention timing affects student learning, we compare prefix intervention (\(\ell_u=0\)) with intervention after a short student attempt (\(\ell_u\in\{512,1024,2048\}\)), keeping the teacher-segment budget fixed.
\Cref{tab:placement} shows that \(\ell_u=512\) middle intervention improves avg@8 and pass@8 over prefix intervention across all three games, with the largest gains on Minesweeper and Kakurasu.
We include a brief case study to further examine why middle intervention helps in practice.

\begingroup
\raggedbottom

\vspace{2pt}
\begin{guidancecasebox}
We compare two explanations for the effect of intervention timing:
\begin{enumerate}[leftmargin=*, nosep, topsep=3pt]
  \item \textls[-15]{\textbf{Later tokens are more informative.}
  Early tokens may largely restate the problem or organize its conditions. Moving supervision later could thus make better use of the same distillation budget.}
  \item \textbf{Student-conditioned guidance is more learnable.}
  Conditioning teacher guidance on the student’s partial solution may make it easier for the student to learn how to continue or recover.
\end{enumerate}

\setlength{\intextsep}{4pt}
\begin{wrapfigure}{r}{0.56\linewidth}
\centering
\includegraphics[width=\linewidth]{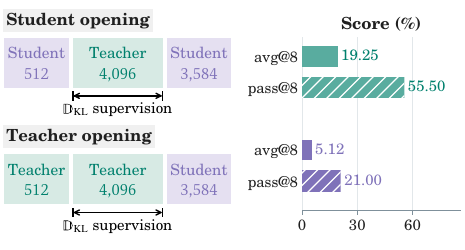}\par
\captionsetup{skip=3pt,font=small}
\caption{Effect of opening source on Minesweeper.}
\label{fig:opening_source}
\end{wrapfigure}
\raggedright
We test these explanations by replacing the student-generated prefix with a teacher-generated one, leaving the rest of the rollout unchanged. Since the teacher may provide a stronger starting point, if middle intervention helps by shifting supervision to more informative tokens, this variant should match or surpass student-prefix intervention.

Yet this replacement substantially reduces performance (\cref{fig:opening_source}), supporting the second explanation.
This result echoes the motivation behind DAgger~\citep{ross2011reduction} and on-policy distillation~\citep{agarwal2024policy}, which expose the learner to states from its own rollouts. From these states, the teacher can help the student build on partial reasoning or correct mistakes.
\end{guidancecasebox}
\vspace{8pt}

A longer student prefix, however, does not necessarily make this guidance more effective.
On Minesweeper, for example, avg@8 falls from 19.25\% at \(\ell_u=512\) to 12.50\% at \(\ell_u=2048\), below the 14.81\% achieved by prefix intervention.
We also observe more truncation with later intervention.
A longer student prefix leaves fewer tokens to continue generating after the teacher segment, especially when earlier steps need to be revised.
As a result, the rollout may reach the length limit before the student can complete the revised reasoning and produce a final answer.
\vspace{14pt}

\begingroup
\setlength{\intextsep}{3pt}
\captionsetup{skip=3pt}
\begin{figure}[H]
\centering
\includegraphics[width=\linewidth]{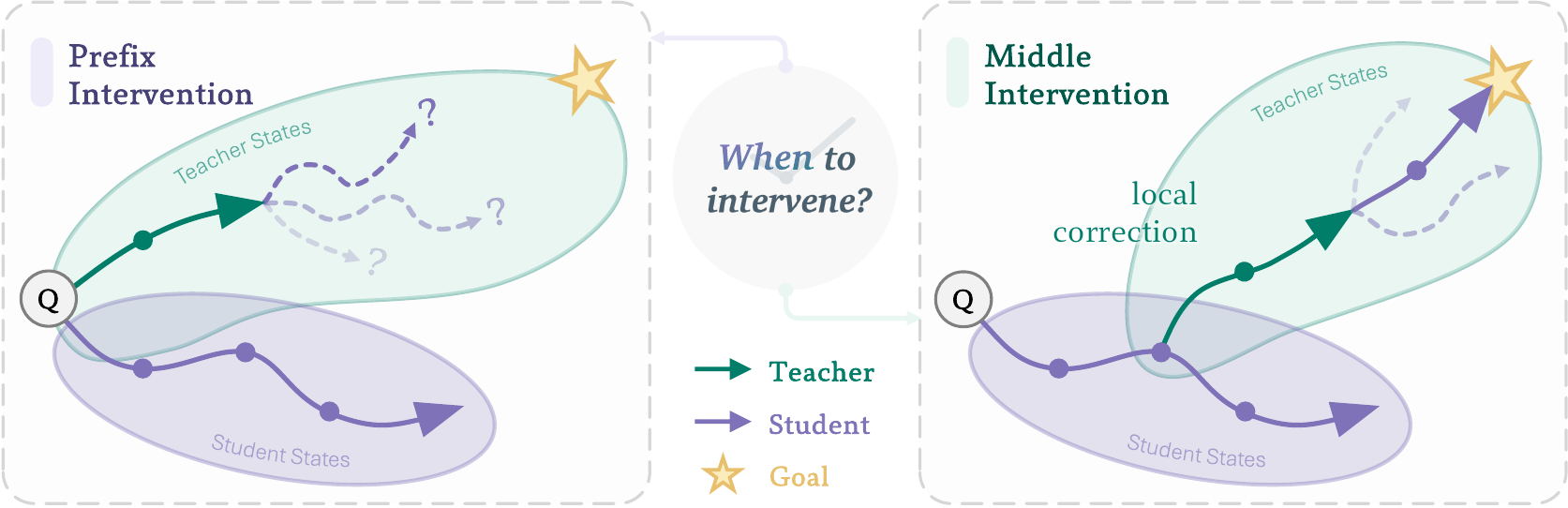}
\caption{
Prefix vs. middle intervention. Guidance is more effective when the teacher can build on the student’s partial solution and correct errors along the way.
}
\label{fig:teacher_intervention}
\end{figure}
\endgroup

\vspace{10pt}
\begin{takeawaybox}
Let the student make a short attempt, then guide it from its own reasoning. 
Intervene early enough to leave room for the student to revise its reasoning and complete the solution.
\end{takeawaybox}

\endgroup

\subsection{How Should the Teacher Supervise?}
\label{sec:migration}

Beyond where the teacher intervenes, the form of supervision also matters for student learning.
A common approach is to minimize the KL divergence between the student and teacher token distributions, providing dense token-level supervision~\citep{agarwal2024policy,xu2025kdrl}.
The forward- and reverse-KL objectives introduced in \cref{sec:preliminaries} both encourage distribution matching, but take expectations under the teacher and student token distributions, respectively.
We compare these objectives within \method to examine how the choice of KL direction affects learning from teacher-generated segments across tasks with different levels of initial competence.

\begin{wrapfigure}{r}{0.48\textwidth}
\centering
\includegraphics[width=\linewidth]{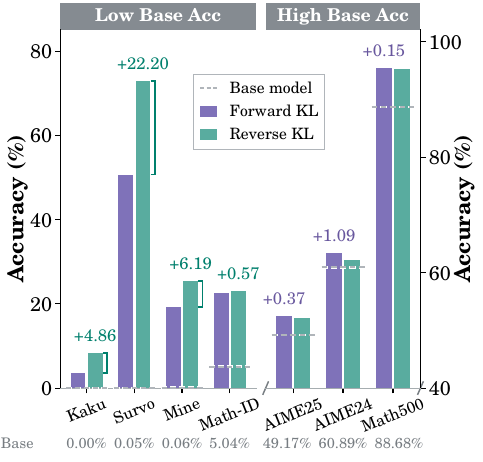}
\captionsetup{skip=4pt}
\caption{Reverse KL favors low initial accuracy; forward KL has a modest advantage at higher accuracy. Dashed lines mark base accuracy. Panels use different scales; labels show gains over the other KL, colored by the better KL.}
\label{fig:kl_comparison}
\end{wrapfigure}

\Cref{fig:kl_comparison} shows that reverse KL performs better on all three hard logic games, where the base student almost never succeeds: mean avg@8 rises from 24.5\% to 35.6\%, and mean pass@8 from 54.2\% to 63.8\%, relative to forward KL (\cref{tab:app_games_full}).
On math benchmarks, both objectives improve average accuracy over the base model.
On the challenging Math-ID, where the base model achieves only 5.04\% accuracy, reverse KL slightly outperforms forward KL by 0.57 percentage points.
On the OOD benchmarks AIME 2024/2025 and MATH500, where the student already solves a substantial fraction of problems, forward KL yields modestly larger gains in average accuracy (\cref{tab:app_math_full}).


A possible explanation lies in the mode-seeking and mode-covering tendencies of reverse and forward KL~\citep{tajwar2024preference,li2026rethinking}.
Reverse KL discourages the student from assigning probability to continuations the teacher considers unlikely, favoring concentration on teacher-preferred modes.
When the student rarely succeeds, learning to reliably produce even one effective reasoning mode could yield substantial progress.
With reliable teacher guidance, this concentration may help the student establish a solution strategy that it previously struggled to produce independently.
Once successful reasoning is already accessible, learning additional effective strategies may offer further room for improvement across different problems.
Forward KL encourages coverage of teacher-supported alternatives, including those the student currently assigns little probability, and may therefore support this broader learning.
The results are consistent with this interpretation, though they do not directly measure changes in reasoning-mode coverage.

We therefore favor reverse KL, given its substantial gains on hard tasks and the modest accuracy advantage of forward KL when initial success rates are higher.

\begin{takeawaybox}
The student's starting point can inform the choice of KL objective.
When success is rare, reverse KL's mode-seeking tendency may help the student establish a viable reasoning strategy.
Forward KL offers a modest accuracy advantage at higher initial success rates, whereas reverse KL yields larger gains on hard tasks.
\end{takeawaybox}

\subsection{What Else Matters for Teacher Guidance?}
\label{sec:teacher_signal_calibration}

\begin{figure}[!t]
\centering
\includegraphics[width=\textwidth]{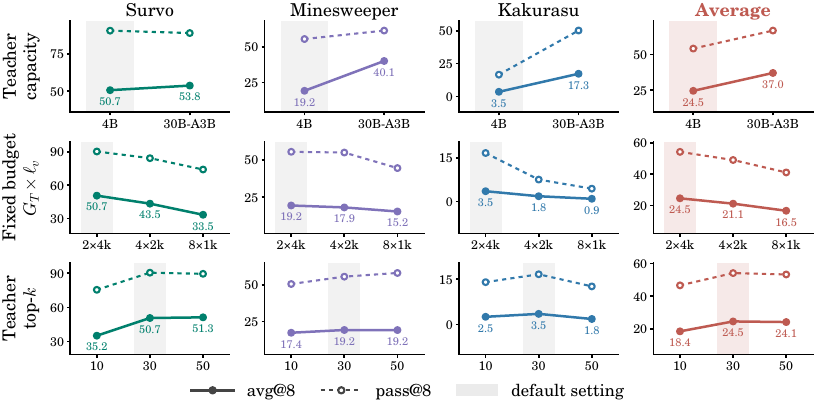}
\captionsetup{skip=3pt}
\caption{
Calibrating teacher guidance on hard logic games under middle intervention with forward KL. Each column varies one factor from the default setting, which is shaded. Solid lines denote avg@8 and dashed lines denote pass@8. Full numbers are given in \cref{tab:app_calibration} in the appendix.
}
\label{fig:calibration}
\end{figure}

With the intervention position and supervision objective fixed, we investigate which other aspects of teacher guidance affect student learning.
We vary teacher capacity, the allocation of a fixed teacher-token budget across assisted rollouts, and the support size of the top-\(k\) approximation.
\Cref{fig:calibration} summarizes ablations, which vary one factor at a time while keeping other settings fixed.

\paragraph{Teacher model capacity.}
We examine whether the student benefits from a stronger teacher by replacing Qwen3-4B-Thinking with Qwen3-30B-A3B-Thinking.
The stronger teacher improves avg@8 across all three games: performance more than doubles on Minesweeper, from \(19.25\%\) to \(40.12\%\), and increases nearly fivefold on Kakurasu, from \(3.52\%\) to \(17.27\%\), while the improvement on Survo is more modest.
These results indicate that \method can make effective use of a more capable teacher despite the larger teacher--student capacity gap.
The additional capability of the teacher translates into improvements in the student's own problem-solving performance, even on games where learning with the default teacher remains difficult.

\paragraph{Allocation of a fixed teacher budget.}
We examine how a fixed teacher-token budget should be distributed across assisted rollouts.
Keeping \(G_T \times \ell_v\) fixed at \(8\)k tokens, we allocate this budget to either two assisted rollouts with a \(4\)k-token teacher budget each, four with \(2\)k each, or eight with \(1\)k each.
Both avg@8 and pass@8 decline monotonically across all three games as the budget is spread across more rollouts; on Survo, avg@8 falls from \(50.70\%\) to \(43.50\%\) and \(33.50\%\).
This suggests that longer teacher segments give the student more opportunity to learn from a sustained sequence of connected reasoning or corrective steps, while shorter segments may end before the teacher has fully developed the reasoning needed for the student to continue.
Under the same teacher-token budget, fewer, longer interventions therefore provide more effective guidance in our experiments.

\Needspace{9\baselineskip}
\paragraph{Top-\(k\) approximation.}
The top-\(k\) approximation controls how much of the teacher's token distribution is available for supervision.
Our results suggest that effective guidance requires a sufficiently broad set of teacher-supported alternatives, but does not necessarily benefit from retaining more of the distribution.
Moving from \(k=10\) to \(k=30\) improves both metrics across all three games, whereas increasing \(k\) to 50 offers no consistent benefit and reduces performance on Kakurasu.
This pattern suggests that the most probable teacher tokens alone may provide too narrow a learning signal, while extending supervision to lower-probability alternatives is not uniformly useful.
We therefore use \(k=30\) as a balance between supervision quality and the cost of storing and matching teacher logits.

\begin{takeawaybox}
A stronger teacher provides better guidance, but making effective use of that guidance also requires careful allocation.
With a fixed teacher-token budget, fewer, longer interventions are more effective than many short ones.
For KL supervision, the teacher distribution should cover enough alternatives for learning; extending to lower-probability tokens offers no consistent benefit.
\end{takeawaybox}

\subsection{Putting the Recipe into Practice}\label{sec:main_results}

Following the recipe above, we configure \method to intervene after a 512-token student prefix, allowing the teacher to build on the student's initial reasoning.
For each prompt, we sample six student-only and two teacher-assisted rollouts, concentrating the teacher-token budget on fewer, longer interventions.
We apply teacher--student KL supervision to the teacher-generated segments with a top-\(k\) support of 30.
We favor reverse KL for its substantial gains when initial success is rare, while also reporting the forward-KL variant.
\Cref{tab:pivot_main} compares these configurations with baselines.

\begin{table}[!htbp]
\definecolor{GainBase}{HTML}{00806D}
\colorlet{GainBase}{GainBase!38!white}
\definecolor{DropBase}{HTML}{BE5A51}
\renewcommand{\gainN}[2]{\colorbox{GainBase!#1!white}{\fontsize{8}{9.5}\selectfont\makebox[2.9em][r]{\strut +#2}}}
\renewcommand{\dropN}[2]{\colorbox{DropBase!#1!white}{\fontsize{8}{9.5}\selectfont\makebox[2.9em][r]{\strut -#2}}}
\centering
\fontsize{8}{9.5}\selectfont
\setlength{\tabcolsep}{1pt}
\setlength{\fboxsep}{1pt}
\renewcommand{\arraystretch}{1.32}
\begin{tabularx}{\textwidth}{l*{7}{>{\centering\arraybackslash}X}}
\toprule
\multirow{2}{*}{Method}
& \multicolumn{3}{c}{Hard Logic Games (Qwen3-1.7B)}
& \multicolumn{4}{c}{Math Benchmarks (Qwen3-4B)} \\
\cmidrule(lr){2-4} \cmidrule(lr){5-8}
& Survo & Minesweeper & Kakurasu & Math-ID & AIME 2024 & AIME 2025 & MATH500 \\
\midrule

Base model
& \makebox[2.35em][r]{0.05}\hspace{0.2em}\phantom{\gainN{0}{0.00}}
& \makebox[2.35em][r]{0.06}\hspace{0.2em}\phantom{\gainN{0}{0.00}}
& \makebox[2.35em][r]{0.00}\hspace{0.2em}\phantom{\gainN{0}{0.00}}
& \makebox[2.35em][r]{5.04}\hspace{0.2em}\phantom{\gainN{0}{0.00}}
& \makebox[2.35em][r]{60.89}\hspace{0.2em}\phantom{\gainN{0}{0.00}}
& \makebox[2.35em][r]{49.17}\hspace{0.2em}\phantom{\gainN{0}{0.00}}
& \makebox[2.35em][r]{88.68}\hspace{0.2em}\phantom{\gainN{0}{0.00}} \\

\midrule
\rowcolor{gray!6}\multicolumn{8}{l}{Baselines} \\
\hspace{0.3em} RL only
& \makebox[2.35em][r]{0.05}\hspace{0.2em}\phantom{\gainN{0}{0.00}}
& \makebox[2.35em][r]{1.94}\hspace{0.2em}\gainN{15}{1.88}
& \makebox[2.35em][r]{0.00}\hspace{0.2em}\phantom{\gainN{0}{0.00}}
& \makebox[2.35em][r]{20.70}\hspace{0.2em}\gainN{60}{15.66}
& \makebox[2.35em][r]{60.62}\hspace{0.2em}\dropN{10}{0.27}
& \makebox[2.35em][r]{49.74}\hspace{0.2em}\gainN{18}{0.57}
& \makebox[2.35em][r]{94.30}\hspace{0.2em}\gainN{58}{5.62} \\

\hspace{0.3em} SFT+RL
& \makebox[2.35em][r]{39.77}\hspace{0.2em}\gainN{38}{39.72}
& \makebox[2.35em][r]{13.44}\hspace{0.2em}\gainN{37}{13.38}
& \makebox[2.35em][r]{1.38}\hspace{0.2em}\gainN{12}{1.38}
& \makebox[2.35em][r]{8.82}\hspace{0.2em}\gainN{15}{3.78}
& \makebox[2.35em][r]{60.94}\hspace{0.2em}\gainN{10}{0.05}
& \makebox[2.35em][r]{49.21}\hspace{0.2em}\gainN{10}{0.04}
& \makebox[2.35em][r]{89.98}\hspace{0.2em}\gainN{20}{1.30} \\

\hspace{0.3em} LUFFY
& \makebox[2.35em][r]{1.30}\hspace{0.2em}\gainN{10}{1.25}
& \makebox[2.35em][r]{0.81}\hspace{0.2em}\gainN{10}{0.75}
& \makebox[2.35em][r]{0.00}\hspace{0.2em}\phantom{\gainN{0}{0.00}}
& \makebox[2.35em][r]{16.66}\hspace{0.2em}\gainN{45}{11.62}
& \makebox[2.35em][r]{60.68}\hspace{0.2em}\dropN{10}{0.21}
& \makebox[2.35em][r]{46.93}\hspace{0.2em}\dropN{35}{2.24}
& \makebox[2.35em][r]{90.60}\hspace{0.2em}\gainN{28}{1.92} \\

\hspace{0.3em} POPE
& \makebox[2.35em][r]{0.03}\hspace{0.2em}\dropN{10}{0.02}
& \makebox[2.35em][r]{0.00}\hspace{0.2em}\dropN{10}{0.06}
& \makebox[2.35em][r]{0.00}\hspace{0.2em}\phantom{\gainN{0}{0.00}}
& \makebox[2.35em][r]{22.09}\hspace{0.2em}\gainN{66}{17.05}
& \makebox[2.35em][r]{60.00}\hspace{0.2em}\dropN{22}{0.89}
& \makebox[2.35em][r]{46.61}\hspace{0.2em}\dropN{40}{2.56}
& \makebox[2.35em][r]{94.20}\hspace{0.2em}\gainN{58}{5.52} \\

\hspace{0.3em} KDRL
& \makebox[2.35em][r]{41.68}\hspace{0.2em}\gainN{40}{41.63}
& \makebox[2.35em][r]{13.88}\hspace{0.2em}\gainN{38}{13.82}
& \makebox[2.35em][r]{2.93}\hspace{0.2em}\gainN{24}{2.93}
& \makebox[2.35em][r]{17.11}\hspace{0.2em}\gainN{47}{12.07}
& \makebox[2.35em][r]{49.11}\hspace{0.2em}\dropN{70}{11.78}
& \makebox[2.35em][r]{43.39}\hspace{0.2em}\dropN{55}{5.78}
& \makebox[2.35em][r]{92.92}\hspace{0.2em}\gainN{45}{4.24} \\

\midrule
\rowcolor{gray!6}\multicolumn{8}{l}{\textbf{\method}} \\
\rowcolor{pivotPurpleLight}
\hspace{0.3em} Forward KL
& \makebox[2.35em][r]{50.70}\hspace{0.2em}\gainN{49}{50.65}
& \makebox[2.35em][r]{19.25}\hspace{0.2em}\gainN{53}{19.19}
& \makebox[2.35em][r]{3.52}\hspace{0.2em}\gainN{29}{3.52}
& \makebox[2.35em][r]{22.48}\hspace{0.2em}\gainN{68}{17.44}
& \makebox[2.35em][r]{\textbf{63.33}}\hspace{0.2em}\gainN{70}{2.44}
& \makebox[2.35em][r]{\textbf{52.45}}\hspace{0.2em}\gainN{70}{3.28}
& \makebox[2.35em][r]{\textbf{95.40}}\hspace{0.2em}\gainN{70}{6.72} \\

\rowcolor{pivotPurpleLight}
\hspace{0.3em} Reverse KL
& \makebox[2.35em][r]{\textbf{72.90}}\hspace{0.2em}\gainN{70}{72.85}
& \makebox[2.35em][r]{\textbf{25.44}}\hspace{0.2em}\gainN{70}{25.38}
& \makebox[2.35em][r]{\textbf{8.38}}\hspace{0.2em}\gainN{70}{8.38}
& \makebox[2.35em][r]{\textbf{23.05}}\hspace{0.2em}\gainN{70}{18.01}
& \makebox[2.35em][r]{62.24}\hspace{0.2em}\gainN{39}{1.35}
& \makebox[2.35em][r]{52.08}\hspace{0.2em}\gainN{62}{2.91}
& \makebox[2.35em][r]{95.25}\hspace{0.2em}\gainN{68}{6.57} \\

\bottomrule
\end{tabularx}
\caption{Main results on hard logic games (Qwen3-1.7B) and math (Qwen3-4B). Math-ID is in-distribution; other math benchmarks are OOD. The base model uses thinking mode. Metrics are avg@8, or avg@64 for AIME. Colored boxes show absolute changes over the base model.}
\label{tab:pivot_main}
\end{table}

On the hard logic games, the base student rarely produces a correct solution, and training with outcome rewards alone does little to improve its performance.
Adding teacher-generated segments to rollouts, as in LUFFY and POPE, brings only limited improvements, with both methods performing worse than even the simple SFT+RL baseline.
KDRL provides token-level supervision along student rollouts and makes greater progress, achieving the strongest baseline performance on all three games.
\method goes further by pairing teacher-generated reasoning segments with direct supervision on those segments: with reverse KL, it raises mean avg@8 from KDRL's \(19.5\%\) to \(35.6\%\).
A similar advantage appears on the hard math split, where the base model solves only \(5.04\%\) of problems.
Here, \method reaches \(23.05\%\), exceeding POPE, the strongest baseline, at \(22.09\%\).

On AIME 2024/2025 and MATH500, where the base student already solves a substantial fraction of problems, teacher guidance does not consistently improve performance.
LUFFY, POPE, and KDRL improve on MATH500 yet fall below even the base model on both AIME benchmarks, while SFT+RL brings little improvement on either.
\method outperforms baselines across all three benchmarks with either KL objective, with forward KL achieving the highest accuracy.
\method therefore benefits students both when initial success is rare and when reasoning abilities are already present.



\FloatBarrier

\section{Conclusion}
Effective teacher guidance in RLVR should support successful reasoning while remaining learnable for the student.
In this work, we identify two complementary mechanisms for achieving this goal: \emph{trajectory-level intervention} and \emph{token-level supervision}.
Existing methods typically emphasize one of them, but our analysis shows that either alone can be insufficient for student learning.
Building on this insight, we introduce \method, which integrates teacher corrections into student rollouts and applies token-level supervision to those corrections.
We examine where to intervene, how to supervise, and how to allocate teacher guidance, yielding a recipe for student learning.
Experiments on logic games and mathematical reasoning demonstrate the effectiveness of combining these mechanisms.


\paragraph{Limitations.}
Our study has several limitations.
First, our evaluation focuses on logic games and mathematical reasoning; applicability to code generation, instruction following, and open-ended reasoning remains to be explored.
Second, \method{} uses fixed intervention and supervision settings throughout training, even as the student's capabilities evolve.
How to effectively adapt teacher guidance to the student's changing needs remains an open question for future work.


\section*{AI use statement}
We used generative AI tools to assist with coding and with drafting, revising, and polishing the manuscript.
The authors reviewed AI-assisted code and text, verified the scientific claims and citations, and remain responsible for the research design, failure analysis, and interpretation of results.
No human-subject data were collected.
We take full responsibility for the manuscript, reported results, and accompanying artifacts, including AI-assisted content.

\section*{Reproducibility Statement}

We describe dataset construction, student and teacher models, training schedules, optimization settings, teacher-intervention budgets, KL-supervision settings, and evaluation protocols in Appendix~\ref{appx:experimental_details}.
Full experimental results are provided in Appendix~\ref{appx:additional_results}.
After the review process concludes, we will open-source the complete training code and publicly release the \method{} model checkpoints to support reproduction of our results and further research on teacher-guided reinforcement learning with verifiable rewards.

\FloatBarrier
\bibliography{iclr2027_conference}
\bibliographystyle{iclr2027_conference}
\clearpage
\appendix

\renewcommand{\floatpagefraction}{0.85}
\renewcommand{\dblfloatpagefraction}{0.85}
\makeatletter
\setlength{\@fptop}{0pt}
\setlength{\@fpsep}{12pt}
\setlength{\@fpbot}{0pt plus 1fil}
\setlength{\@dblfptop}{0pt}
\setlength{\@dblfpsep}{12pt}
\setlength{\@dblfpbot}{0pt plus 1fil}
\makeatother

\begingroup
\fontsize{10.5}{12.6}\selectfont
\setlength{\parskip}{4pt}
\section{Experimental Setup}
\label{appx:experimental_details}
We detail the dataset construction, training configuration, evaluation protocol, and baseline implementations.

\paragraph{Dataset construction.}
For synthetic logic games, we follow SynLogic~\citep{liu2026synlogic} and construct three datasets: Survo, Minesweeper, and Kakurasu. 
Each game is generated at a hard difficulty setting, controlled by grid size, the number of unknown entries, and constraint density, at which the base student is near zero accuracy. Each game has 2k training problems and a held-out test split of 500 problems.
For mathematical reasoning, we pool problems from Polaris-53K~\citep{an2025polaris} and the math subset of Skywork-OR1-RL-Data~\citep{he2025skywork}, and screen them with the base student: Qwen3-4B samples 8 responses per problem, and we keep problems solved at most 3 times. 
After removing problems with invalid gold answers and decontaminating against AIME 2024/2025 and MATH500, we sample 30k problems for training, 1k for validation, and a held-out in-distribution test split of 2k problems, denoted Math-ID; 77.7\% of the training problems are never solved by the base student.

\paragraph{Training setup.}
All experiments are conducted using the verl framework~\citep{sheng2025hybridflow}. 
We use GRPO~\citep{shao2024deepseekmath} as the policy optimization algorithm, with reference KL coefficient $\beta_{\mathrm{ref}}=0.01$ and clipping coefficient $\varepsilon=0.2$. 
Unless otherwise specified, we use a learning rate of $3\times10^{-6}$, a rollout batch size of 64 prompts, a mini-batch size of 32, 8 rollouts per prompt, a maximum prompt length of 2,048 tokens, and sampling temperature 0.7 for both student and teacher. 
Logic-game experiments use Qwen3-1.7B as the student and Qwen3-4B-Thinking as the teacher~\citep{yang2025qwen3}; we train for 3 epochs with a maximum response length of 8,192 tokens and compute rewards with the SynLogic verifier. 
Math experiments use Qwen3-4B as the student and the same teacher; we train for 1 epoch with a maximum response length of 16,384 tokens and compute rewards with Math-Verify. 
No additional format or length rewards are used.

\paragraph{PIVOT configuration.}
For each prompt, we sample $G_S+G_T=8$ rollouts, where $G_S=6$ are student-only rollouts and $G_T=2$ are teacher-assisted rollouts. 
By default, the teacher intervenes after a student prefix of $\ell_u=512$ tokens and generates at most $\ell_v=4{,}096$ tokens for logic games and $\ell_v=8{,}192$ tokens for math. 
The student then resumes autoregressive generation from the teacher-augmented context. 
The teacher-generated segment is supervised with a token-level KL objective toward the teacher distribution; the placement and calibration ablations use forward KL, and we compare both directions in Section~\ref{sec:migration}. 
The KL is computed on the teacher's top-30 tokens, with the remaining probability mass collected into a single tail bucket. 
The teacher KL coefficient $\beta_{\mathrm{T}}$ is cosine-decayed from 0.1 to 0. 
Ablations on the intervention position $\ell_u$, the teacher model, the allocation of the teacher-token budget between $G_T$ and $\ell_v$, and the distillation top-$k$ are provided in Sections~\ref{sec:learnability} and~\ref{sec:teacher_signal_calibration}, with full numbers in Tables~\ref{tab:placement} and~\ref{tab:app_calibration}.

\paragraph{Evaluation.}
For synthetic logic games, we evaluate models on the held-out hard splits and report avg@8 and pass@8. 
For mathematical reasoning, we evaluate in distribution on Math-ID and out of distribution on MATH500~\citep{hendrycks2021measuring} and AIME 2024/2025~\citep{li2024numinamath}, reporting avg@8 and pass@8 for Math-ID and MATH500 and avg@64 and pass@64 for AIME 2024/2025. 
Decoding uses temperature 0.7 and the training response-length limit.

\paragraph{Baseline methods.}
We compare \method{} with five representative baselines that address sparse rewards in different ways. 
All baselines share the GRPO setup above and the budget of 8 rollouts per prompt. 
RL only is the standard on-policy RLVR baseline without a teacher. 
SFT+RL first fine-tunes the student on 5k verified-correct teacher-generated traces from the same training distribution and then continues RLVR training. 
LUFFY~\citep{yan2025learning} replaces one of the 8 rollouts with a full teacher-generated trajectory and trains on it with LUFFY's policy-shaping objective ($\gamma=0.1$). 
POPE~\citep{qu2026pope} starts 2 of the 8 rollouts from a teacher-generated prefix whose length matches our default segment length $\ell_v$ (4,096 tokens for logic games, 8,192 for math); prefix tokens are excluded from all losses and the student completes the rollout. 
KDRL~\citep{xu2025kdrl} adds no teacher tokens to the trajectory; it combines GRPO with a reverse-KL distillation term on every student-generated token, computed on the student's top-30 tokens with a constant coefficient of 0.1.
\endgroup

\newcommand{\appResultN}[1]{\makebox[2.55em][r]{#1}\hspace{0.45em}\makebox[3.05em][r]{}}
\newcommand{\appResultG}[3]{\makebox[2.55em][r]{#1}\hspace{0.45em}\makebox[3.05em][r]{{\fontsize{7.5}{9}\selectfont\textcolor{figteal!75!black}{+#3}}}}
\newcommand{\appResultD}[3]{\makebox[2.55em][r]{#1}\hspace{0.45em}\makebox[3.05em][r]{{\fontsize{7.5}{9}\selectfont\textcolor{red!45!black}{-#3}}}}
\newcommand{\appDefaultMark}{\,\textcolor{pivotPurpleDark}{\textsuperscript{\ensuremath{\ast}}}}

\clearpage
\section{Additional Experimental Results}
\label{appx:additional_results}
We complement the main results with a prefix-only comparison, complete benchmark results, and teacher-guidance ablations. In the benchmark tables, small signed values give absolute percentage-point changes from the base model; bold values identify the best result in each column.

\subsection{Prefix-Only Guidance}
Table~\ref{tab:prefix_only_survo} compares POPE-style prefix-only guidance with supervised prefix training on Survo. Oracle prefixes are verifier-filtered teacher rollouts. They improve on noisy prefixes, but both prefix-only variants remain below 1\% on avg@8 and pass@8. Prefix-Fwd reaches 59.05 avg@8 and 94.20 pass@8, suggesting that teacher prefixes provide a stronger learning signal when they are directly supervised rather than supplied only as context.

\begin{table}[H]
\centering
\small
\setlength{\tabcolsep}{7pt}
\renewcommand{\arraystretch}{1.12}
\begin{tabularx}{0.72\linewidth}{Xcc}
\toprule
Method & avg@8 & pass@8 \\
\midrule
POPE w/ noisy prefix & 0.03 & 0.20 \\
POPE w/ oracle prefix & 0.14 & 0.91 \\
\midrule
\rowcolor{pivotPurpleLight}
Prefix-Fwd & \textbf{59.05} & \textbf{94.20} \\
\bottomrule
\end{tabularx}
\caption{Prefix-only guidance and supervised prefix training on Survo. Oracle prefixes use verifier-filtered teacher rollouts.}
\label{tab:prefix_only_survo}
\end{table}

\FloatBarrier
\Needspace{23\baselineskip}
\subsection{Hard Logic Games}
Table~\ref{tab:app_games_full} reports avg@8 and pass@8 on the held-out hard splits of Survo, Minesweeper, and Kakurasu. Both \method{} variants outperform every baseline on all six metrics. Reverse KL achieves the highest scores throughout, with the largest absolute gain on Survo: 72.85 percentage points in avg@8 and 97.00 points in pass@8 over the base model. Kakurasu remains the most challenging game, with a best avg@8 of 8.38 despite these gains.

\begin{table}[H]
\centering
\fontsize{9}{11}\selectfont
\setlength{\tabcolsep}{2pt}
\renewcommand{\arraystretch}{1.13}
\begin{tabularx}{\linewidth}{l*{6}{>{\centering\arraybackslash}X}}
\toprule
\multirow{2}{*}{Method}
& \multicolumn{2}{c}{Survo}
& \multicolumn{2}{c}{Minesweeper}
& \multicolumn{2}{c}{Kakurasu} \\
\cmidrule(lr){2-3} \cmidrule(lr){4-5} \cmidrule(lr){6-7}
& avg@8 & pass@8 & avg@8 & pass@8 & avg@8 & pass@8 \\
\midrule
Base model
& \appResultN{0.05} & \appResultN{0.40}
& \appResultN{0.06} & \appResultN{0.50}
& \appResultN{0.00} & \appResultN{0.00} \\

\midrule
\rowcolor{gray!6}\multicolumn{7}{l}{Baselines} \\
\hspace{0.3em} RL only
& \appResultN{0.05} & \appResultN{0.40}
& \appResultG{1.94}{15}{1.88} & \appResultG{8.00}{18}{7.50}
& \appResultN{0.00} & \appResultN{0.00} \\

\hspace{0.3em} SFT+RL
& \appResultG{39.77}{38}{39.72} & \appResultG{86.40}{62}{86.00}
& \appResultG{13.44}{37}{13.38} & \appResultG{41.50}{46}{41.00}
& \appResultG{1.38}{12}{1.38} & \appResultG{7.60}{17}{7.60} \\

\hspace{0.3em} LUFFY
& \appResultG{1.30}{10}{1.25} & \appResultG{9.00}{12}{8.60}
& \appResultG{0.81}{10}{0.75} & \appResultG{5.00}{12}{4.50}
& \appResultN{0.00} & \appResultN{0.00} \\

\hspace{0.3em} POPE
& \appResultD{0.03}{10}{0.02} & \appResultD{0.20}{10}{0.20}
& \appResultD{0.00}{10}{0.06} & \appResultD{0.00}{10}{0.50}
& \appResultN{0.00} & \appResultN{0.00} \\

\hspace{0.3em} KDRL
& \appResultG{41.68}{40}{41.63} & \appResultG{81.40}{58}{81.00}
& \appResultG{13.88}{38}{13.82} & \appResultG{42.50}{47}{42.00}
& \appResultG{2.93}{24}{2.93} & \appResultG{12.40}{28}{12.40} \\

\midrule
\rowcolor{gray!6}\multicolumn{7}{l}{\textbf{\method}} \\
\rowcolor{pivotPurpleLight}
\hspace{0.3em} Forward KL
& \appResultG{50.70}{49}{50.65} & \appResultG{90.40}{65}{90.00}
& \appResultG{19.25}{53}{19.19} & \appResultG{55.50}{62}{55.00}
& \appResultG{3.52}{29}{3.52} & \appResultG{16.60}{37}{16.60} \\

\rowcolor{pivotPurpleLight}
\hspace{0.3em} Reverse KL
& \appResultG{\textbf{72.90}}{70}{72.85} & \appResultG{\textbf{97.40}}{70}{97.00}
& \appResultG{\textbf{25.44}}{70}{25.38} & \appResultG{\textbf{63.00}}{70}{62.50}
& \appResultG{\textbf{8.38}}{70}{8.38} & \appResultG{\textbf{31.00}}{70}{31.00} \\

\bottomrule
\end{tabularx}
\caption{Full results on hard logic games (Qwen3-1.7B; base model in thinking mode). Signed values show changes from the base model. Shading identifies \method{}; bold indicates the best result in each column.}
\label{tab:app_games_full}
\end{table}

\FloatBarrier
\Needspace{43\baselineskip}
\subsection{Mathematical Reasoning}
Table~\ref{tab:app_math_full} separates the full math results by evaluation budget: avg@8/pass@8 for Math-ID and MATH500, and avg@64/pass@64 for AIME 2024/2025. Reverse KL performs best on the held-out in-distribution split (Math-ID), while forward KL achieves the highest scores on all three out-of-distribution benchmarks. Both \method{} variants improve avg@$k$ over the base model on every benchmark, with no decrease in pass@$k$.

\begin{table}[H]
\centering
\fontsize{10}{12.2}\selectfont
\setlength{\tabcolsep}{4pt}
\renewcommand{\arraystretch}{1.15}
{\small\bfseries\color{pivotPurpleDark} Math-ID and MATH500\par}
\vspace{4pt}
\begin{tabularx}{\linewidth}{l*{4}{>{\centering\arraybackslash}X}}
\toprule
\multirow{2}{*}{Method}
& \multicolumn{2}{c}{Math-ID}
& \multicolumn{2}{c}{MATH500} \\
\cmidrule(lr){2-3} \cmidrule(lr){4-5}
& avg@8 & pass@8 & avg@8 & pass@8 \\
\midrule
Base model
& \appResultN{5.04} & \appResultN{22.20} & \appResultN{88.68} & \appResultN{92.20} \\
\midrule
\rowcolor{gray!6}\multicolumn{5}{l}{Baselines} \\
\hspace{0.3em} RL only
& \appResultG{20.70}{60}{15.66} & \appResultG{37.00}{50}{14.80} & \appResultG{94.30}{58}{5.62} & \appResultG{98.00}{62}{5.80} \\
\hspace{0.3em} SFT+RL
& \appResultG{8.82}{15}{3.78} & \appResultG{28.20}{20}{6.00} & \appResultG{89.98}{20}{1.30} & \appResultG{94.20}{22}{2.00} \\
\hspace{0.3em} LUFFY
& \appResultG{16.66}{45}{11.62} & \appResultG{33.00}{37}{10.80} & \appResultG{90.60}{28}{1.92} & \appResultG{97.20}{54}{5.00} \\
\hspace{0.3em} POPE
& \appResultG{22.09}{66}{17.05} & \appResultG{38.90}{57}{16.70} & \appResultG{94.20}{58}{5.52} & \appResultG{97.60}{58}{5.40} \\
\hspace{0.3em} KDRL
& \appResultG{17.11}{47}{12.07} & \appResultG{30.20}{27}{8.00} & \appResultG{92.92}{45}{4.24} & \appResultG{96.80}{48}{4.60} \\
\midrule
\rowcolor{gray!6}\multicolumn{5}{l}{\textbf{\method}} \\
\rowcolor{pivotPurpleLight}
\hspace{0.3em} Forward KL
& \appResultG{22.48}{68}{17.44} & \appResultG{41.70}{66}{19.50} & \appResultG{\textbf{95.40}}{70}{6.72} & \appResultG{\textbf{98.80}}{70}{6.60} \\
\rowcolor{pivotPurpleLight}
\hspace{0.3em} Reverse KL
& \appResultG{\textbf{23.05}}{70}{18.01} & \appResultG{\textbf{42.75}}{70}{20.55} & \appResultG{95.25}{68}{6.57} & \appResultG{98.60}{68}{6.40} \\
\bottomrule
\end{tabularx}

\vspace{10pt}
{\small\bfseries\color{pivotPurpleDark} AIME 2024 and AIME 2025\par}
\vspace{4pt}
\begin{tabularx}{\linewidth}{l*{4}{>{\centering\arraybackslash}X}}
\toprule
\multirow{2}{*}{Method}
& \multicolumn{2}{c}{AIME 2024}
& \multicolumn{2}{c}{AIME 2025} \\
\cmidrule(lr){2-3} \cmidrule(lr){4-5}
& avg@64 & pass@64 & avg@64 & pass@64 \\
\midrule
Base model
& \appResultN{60.89} & \appResultN{83.33} & \appResultN{49.17} & \appResultN{76.67} \\
\midrule
\rowcolor{gray!6}\multicolumn{5}{l}{Baselines} \\
\hspace{0.3em} RL only
& \appResultD{60.62}{10}{0.27} & \appResultN{83.33} & \appResultG{49.74}{18}{0.57} & \appResultN{76.67} \\
\hspace{0.3em} SFT+RL
& \appResultG{60.94}{10}{0.05} & \appResultN{83.33} & \appResultG{49.21}{10}{0.04} & \appResultN{76.67} \\
\hspace{0.3em} LUFFY
& \appResultD{60.68}{10}{0.21} & \appResultN{83.33} & \appResultD{46.93}{35}{2.24} & \appResultN{76.67} \\
\hspace{0.3em} POPE
& \appResultD{60.00}{22}{0.89} & \appResultN{83.33} & \appResultD{46.61}{40}{2.56} & \appResultN{76.67} \\
\hspace{0.3em} KDRL
& \appResultD{49.11}{70}{11.78} & \appResultD{76.67}{58}{6.66} & \appResultD{43.39}{55}{5.78} & \appResultD{63.33}{70}{13.34} \\
\midrule
\rowcolor{gray!6}\multicolumn{5}{l}{\textbf{\method}} \\
\rowcolor{pivotPurpleLight}
\hspace{0.3em} Forward KL
& \appResultG{\textbf{63.33}}{70}{2.44} & \appResultG{\textbf{86.67}}{70}{3.34} & \appResultG{\textbf{52.45}}{70}{3.28} & \appResultG{\textbf{80.00}}{70}{3.33} \\
\rowcolor{pivotPurpleLight}
\hspace{0.3em} Reverse KL
& \appResultG{62.24}{39}{1.35} & \appResultN{83.33} & \appResultG{52.08}{62}{2.91} & \appResultN{76.67} \\
\bottomrule
\end{tabularx}
\caption{Full mathematical reasoning results (Qwen3-4B, thinking mode). Signed values show changes from the base model. The two panels use different sampling budgets; bold identifies the best result for each benchmark and metric.}
\label{tab:app_math_full}
\end{table}

\FloatBarrier
\Needspace{34\baselineskip}
\subsection{Teacher-Guidance Ablations}
Table~\ref{tab:app_calibration} gives the complete results behind Figure~\ref{fig:calibration}. Each block changes one factor while retaining middle intervention and forward KL; the default setting is shaded and repeated for reference. The prefix-length ablation appears in Table~\ref{tab:placement}.

A larger teacher improves avg@8 on all three games, although Survo pass@8 decreases slightly. At a fixed teacher-token budget, two longer segments outperform four or eight shorter segments on every metric. For the teacher's top-$k$ distribution, the default $k=30$ leads on four of the six metrics; $k=50$ gives the highest Survo avg@8 and Minesweeper pass@8.

\begin{table}[H]
\centering
\fontsize{10}{12.2}\selectfont
\setlength{\tabcolsep}{3pt}
\renewcommand{\arraystretch}{1.25}
\begin{tabularx}{\textwidth}{l*{6}{>{\centering\arraybackslash}X}}
\toprule
\multirow{2}{*}{Setting}
& \multicolumn{2}{c}{Survo}
& \multicolumn{2}{c}{Minesweeper}
& \multicolumn{2}{c}{Kakurasu} \\
\cmidrule(lr){2-3} \cmidrule(lr){4-5} \cmidrule(lr){6-7}
& avg@8 & pass@8 & avg@8 & pass@8 & avg@8 & pass@8 \\
\midrule
\rowcolor{gray!6}\multicolumn{7}{l}{\textit{Teacher capacity}} \\
\rowcolor{pivotPurpleLight}
\hspace{0.3em} Qwen3-4B-Thinking\appDefaultMark
& 50.70 & \textbf{90.40} & 19.25 & 55.50 & 3.52 & 16.60 \\
\hspace{0.3em} Qwen3-30B-A3B-Thinking
& \textbf{53.75} & 88.80 & \textbf{40.12} & \textbf{61.40} & \textbf{17.27} & \textbf{50.20} \\
\midrule
\rowcolor{gray!6}\multicolumn{7}{l}{\textit{Fixed teacher budget} ($G_T\times\ell_v=8$k)} \\
\rowcolor{pivotPurpleLight}
\hspace{0.3em} $2\times4$k\appDefaultMark
& \textbf{50.70} & \textbf{90.40} & \textbf{19.25} & \textbf{55.50} & \textbf{3.52} & \textbf{16.60} \\
\hspace{0.3em} $4\times2$k
& 43.50 & 84.40 & 17.94 & 55.00 & 1.80 & 7.50 \\
\hspace{0.3em} $8\times1$k
& 33.50 & 74.20 & 15.19 & 44.50 & 0.93 & 4.40 \\
\midrule
\rowcolor{gray!6}\multicolumn{7}{l}{\textit{Teacher top-$k$}} \\
\hspace{0.3em} $k=10$
& 35.23 & 75.40 & 17.38 & 50.50 & 2.55 & 14.00 \\
\rowcolor{pivotPurpleLight}
\hspace{0.3em} $k=30$\appDefaultMark
& 50.70 & \textbf{90.40} & \textbf{19.25} & 55.50 & \textbf{3.52} & \textbf{16.60} \\
\hspace{0.3em} $k=50$
& \textbf{51.27} & 89.40 & 19.19 & \textbf{58.00} & 1.83 & 12.60 \\
\bottomrule
\end{tabularx}
\caption{Teacher-guidance ablations with middle intervention and forward KL. Shaded rows repeat the default setting (${}^\ast$). Bold values identify the best result within each block.}
\label{tab:app_calibration}
\end{table}
\FloatBarrier

\clearpage
\section{Case Studies on Intervention Placement}
\label{appx:case_placement}

\newcommand{\ctk}[1]{{\fontsize{8}{10}\selectfont\ttfamily\textcolor{gray}{[#1]}}}
\newcommand{\cmk}[1]{\textcolor{pivotPurpleDark}{\textbf{#1}}}
\newcommand{\cfence}{\textasciigrave\textasciigrave\textasciigrave}
\newcommand{\cq}[1]{{\itshape #1}}
\newcommand{\cnote}[1]{\textbf{#1}}
\newtcolorbox{casebox}[1]{
  enhanced,
  colback=white, colframe=pivotRule,
  colbacktitle=pivotPurpleLight, coltitle=pivotPurpleDark,
  title={#1}, fonttitle=\bfseries\fontsize{10}{12}\selectfont,
  boxrule=0.6pt, arc=1.2mm,
  left=2mm, right=2mm, top=1.5mm, bottom=1.5mm,
  before skip=7pt, after skip=5pt,
  fontupper=\fontsize{9.5}{11.7}\selectfont,
  before upper={\setlength{\parskip}{3pt}}
}
\newenvironment{casetable}{%
  \setlength{\tabcolsep}{0pt}%
  \renewcommand{\arraystretch}{1.04}%
  \arrayrulecolor{pivotRule}%
  \tabularx{\linewidth}{@{}>{\raggedright\arraybackslash}p{.60\linewidth}@{\hspace{12pt}}>{\raggedright\arraybackslash}X@{}}
  \textcolor{gray}{\fontsize{8}{10}\selectfont\textsc{trajectory excerpts}} &
  \textcolor{gray}{\fontsize{8}{10}\selectfont\textsc{observations}}\\[3pt]
}{\endtabularx\arrayrulecolor{black}}
\newcommand{\cseg}[2]{%
  \specialrule{.4pt}{5pt}{3pt}%
  \multicolumn{2}{@{}p{\linewidth}@{}}{\textcolor{#2}{\bfseries #1}}\\[4pt]}
\newcommand{\crow}{\\[6pt]}
\newcommand{\cend}{\\[1pt]}
\newcommand{\casecaption}[2]{\par\noindent\begin{minipage}{\linewidth}\captionsetup{hypcap=false}\captionof{figure}{#1}\label{#2}\end{minipage}\par}

We trace one Survo puzzle through the three intervention positions in Table~\ref{tab:placement} (trained student, step~60).
The comparison asks whether the resulting state can be completed by the student, and whether enough response budget remains to emit the answer.

\begin{tcolorbox}[enhanced,colback=softgray,colframe=pivotRule,boxrule=.5pt,
  arc=1.2mm,left=2.5mm,right=2.5mm,top=2mm,bottom=2mm,
  fontupper=\fontsize{9.5}{11.7}\selectfont,before skip=5pt,after skip=5pt]
\begin{minipage}[t]{.32\linewidth}
\vspace{0pt}\textbf{Shared puzzle}\par\vspace{4pt}
\begingroup\setlength{\tabcolsep}{3.2pt}\renewcommand{\arraystretch}{1.1}
\begin{tabular}{@{}cccccc|c@{}}
8 & 2 & 7 & 8 & \hly{X} & 2 & \textcolor{gray}{32}\\
5 & 5 & 6 & 5 & 7 & 1 & \textcolor{gray}{29}\\
3 & 6 & 7 & \hly{X} & 5 & 9 & \textcolor{gray}{32}\\
2 & 6 & 8 & \hly{X} & \hly{X} & 4 & \textcolor{gray}{31}\\
7 & 3 & \hly{X} & 5 & \hly{X} & 6 & \textcolor{gray}{35}\\
\hly{X} & \hly{X} & 5 & 9 & \hly{X} & \hly{X} & \textcolor{gray}{38}\\\hline
\textcolor{gray}{29} & \textcolor{gray}{30} & \textcolor{gray}{39} & \textcolor{gray}{35} & \textcolor{gray}{34} & \textcolor{gray}{30} & \textcolor{gray}{197}
\end{tabular}\endgroup
\end{minipage}\hfill
\begin{minipage}[t]{.64\linewidth}
\vspace{0pt}\textbf{Prompt and controlled continuations}\par\vspace{3pt}
Fill the 10 cells marked \hly{X} with $[5,8,4,6,2,6,4,8,8,5]$, each used once, so that the last number of every row and column is the sum of the others. Answer with the filled matrix in a Python code block.

For each $\ell_u\!\in\!\{0,512,2048\}$, freeze the rollout from the corresponding training run after the student prefix and the $4{,}096$-token teacher segment. Continue that exact context four times each with the teacher and one fixed student (the $\ell_u=512$ checkpoint, step~60; temperature~0.7). The $8{,}192$-token response limit leaves $4{,}096$, $3{,}584$, or $2{,}048$ continuation tokens, respectively.
\end{minipage}
\end{tcolorbox}
{\fontsize{9}{11}\selectfont\textcolor{gray}{Excerpts follow generation order. Token offsets \ctk{+k} start at the continuation; other offsets start at the response. \hlp{Pink} marks wasted or failing reasoning; \hlg{green} marks progress.}\par}

\begin{casebox}{Case 1\enspace /\enspace Prefix intervention ($\ell_u=0$)}
\begin{casetable}
\cseg{Teacher segment \hfill tokens 0--4,096}{figteal}
\cmk{1}\enspace\ctk{0.5k} \cq{\hlp{I think I have a misunderstanding. Let me read the problem again.}} \ctk{1.0k} \cq{\hlp{I think there's confusion in indexing.}}
& \cmk{1}\enspace\cnote{Slow start.} Re-reads the task for $\sim$1.5k tokens.
\crow
\cmk{2}\enspace\ctk{2.5k} \cq{\hlg{Yes, so row 0, col4 must be 5.}} \ctk{3.0k} \cq{\hlg{So row 2, col3 must be 2.}}
& \cmk{2}\enspace\cnote{2/10 cells solved.} Most cells remain unresolved at the hand-off.
\crow
\cmk{3}\enspace\ctk{3.4k} \cq{\hlp{Let's use different notation.} [\dots] let A = M[3][3], B = M[3][4], so A + B = 11 [\dots] E + F + G + H = 24} \ctk{4.1k} \textcolor{gray}{(segment ends)}
& \cmk{3}\enspace\cnote{Notation changes.} Eight open cells are renamed $A,\dots,H$; the student inherits this system.
\cend
\cseg{Student continuation \hfill budget 4,096}{figpurple}
\cmk{4}\enspace\ctk{+0.5k} \cq{with \hlp{G + H = 24} [\dots] 14 < 24. \hlp{Contradiction.} Wait, this can't be. \hlp{What's wrong?}} \ctk{+2.4k} \cq{\hlp{Contradiction again.}} \ctk{+4.1k} \textcolor{gray}{(length limit, no answer)}
& \cmk{4}\enspace\cnote{Student: 0/4.} Drops $E,F$ and loops over contradictions, or solves the grid but re-checks until truncation.
\cend
\cseg{Teacher continuation from the same hand-off}{figteal}
\cmk{5}\enspace\ctk{+0k} \cq{\hlg{thus E = 4} [\dots] \hlg{thus C = 6} [\dots] All good.} \ctk{+3.7k} \texttt{\cfence python}
& \cmk{5}\enspace\cnote{Teacher: 4/4.} Completes from the same state and with the same remaining budget.
\cend
\end{casetable}
\end{casebox}
\casecaption{Prefix versus middle intervention on one Survo puzzle. With prefix intervention, the hand-off state is completed by the teacher (4/4), but not by the student (0/4). The matched middle intervention continues on the next page.}{fig:case_prefix_middle}

\clearpage
\begin{casebox}{Case 1 (continued)\enspace /\enspace Middle intervention ($\ell_u=512$)}
\begin{casetable}
\cseg{Student prefix \hfill tokens 0--512}{figpurple}
\cmk{1}\enspace\ctk{0.1k} \cq{Row 0: [8, 2, 7, 8, 'X', 2, 32] [\dots] Row 6: [29, 30, 39, 35, 34, 30, 197]} \ctk{0.5k} \cq{\hlg{Let me count.}}
& \cmk{1}\enspace\cnote{Student's set-up.} Lists the rows and states the sum rule, giving the teacher a context built by the student.
\cend
\cseg{Teacher segment \hfill tokens 512--4,608}{figteal}
\cmk{2}\enspace\ctk{0.6k} \cq{8+2+7+8+2 = 27, so 27 + X = 32, \hlg{so X = 5.}}
& \cmk{2}\enspace\cnote{Immediate progress.} First cell solved 87 tokens after take-over, versus 2.5k from scratch; the teacher uses the student's notation.
\crow
\cmk{3}\enspace\ctk{2.9k} \cq{\hlg{E is (5,0), so (5,0) = 4}} \ctk{3.3k} \cq{\hlg{thus C = 6}} \ctk{4.1k} \cq{\hlg{thus H=8}} \ctk{4.3k} \cq{\hlg{Perfect.}} \ctk{4.6k} \textcolor{gray}{(segment ends)}
& \cmk{3}\enspace\cnote{10/10 cells solved.} All cells are fixed and verified at hand-off; 3,584 tokens remain for the response.
\cend
\cseg{Student continuation \hfill budget 3,584}{figpurple}
\cmk{4}\enspace\ctk{+1.4k} \cq{\hlg{All good.}} \texttt{</think> \cfence python}
& \cmk{4}\enspace\cnote{Student and teacher: 4/4.} The student finishes in 1.4k--1.8k tokens.
\cend
\end{casetable}
\end{casebox}
\begin{minipage}{\linewidth}
\captionsetup{type=figure,hypcap=false}\ContinuedFloat
\caption{Prefix versus middle intervention (continued). The student-conditioned middle state is completed by both models. All excerpts and counts refer to the same puzzle and the controlled continuations described on the preceding page.}
\end{minipage}

\subsection*{Comparison at the hand-off}
\begin{tcolorbox}[enhanced,colback=softgray,colframe=pivotRule,boxrule=.5pt,
  arc=1.2mm,left=3mm,right=3mm,top=2mm,bottom=2mm,
  fontupper=\fontsize{10}{12.5}\selectfont]
\setlength{\tabcolsep}{5pt}\renewcommand{\arraystretch}{1.2}
\begin{tabularx}{\linewidth}{@{}>{\raggedright\arraybackslash}Xcc@{}}
& \textbf{Prefix} & \textbf{Middle}\\
\midrule
Student prefix & 0 tokens & 512 tokens\\
Cells solved at hand-off & 2/10 & 10/10\\
Continuation budget & 4,096 tokens & 3,584 tokens\\
Student completions & \textcolor{pivotPurpleDark}{\textbf{0/4}} & \textcolor{pivotPurpleDark}{\textbf{4/4}}\\
Teacher completions & 4/4 & 4/4\\
\end{tabularx}
\end{tcolorbox}

\textbf{Observation.} Both frozen states are solvable for the teacher under the original response limit. Only the middle state is completed by the student, even though it leaves fewer continuation tokens.

\textbf{Interpretation.} In this example, intervening after the student's initial set-up lets the teacher make earlier progress and return a fully resolved grid. The prefix intervention instead returns an incomplete state with new notation that the student struggles to use. These four continuations illustrate the distinction between a state the teacher can solve and a state the student can reliably finish; they do not isolate notation or timing as the sole cause.

\clearpage
\begin{casebox}{Case 2\enspace /\enspace Late intervention ($\ell_u=2048$)}
\begin{casetable}
\cseg{Student prefix \hfill tokens 0--2,048}{figpurple}
\cmk{1}\enspace\ctk{0.5k} \cq{\hlp{So there are 12 empty cells}, but the list says 12? [\dots] total 1+2+2+2+2=9? [\dots] \hlp{= 10 empty cells.}} \ctk{1.8k} \cq{Let me write the equations.}
& \cmk{1}\enspace\cnote{0/10 cells solved.} The student recounts cells and begins writing equations without solving a cell in 2,048 tokens.
\cend
\cseg{Teacher segment \hfill tokens 2,048--6,144}{figteal}
\cmk{2}\enspace\ctk{2.3k} \cq{\hlg{thus X=5.}} \ctk{2.4k} \cq{\hlg{So row2, col3 should be 2.}} \ctk{6.1k} \cq{Let I = (4,2), J = (4,4) = D. So row4: col0=7, col1=3, col2=I, \hlp{col3=}} \textcolor{gray}{(segment cap)}
& \cmk{2}\enspace\cnote{6/10 cells solved.} The teacher makes partial progress before its segment ends. Only 2,048 continuation tokens remain.
\cend
\cseg{Student continuation \hfill budget 2,048}{figpurple}
\cmk{3}\enspace\ctk{+0.6k} \cq{\hlg{D=8, so (4,4)=8} [\dots] \hlg{So (4,2)=6}}
& \cmk{3}\enspace\cnote{Grid solved.} The last four cells are fixed within 0.7k tokens.
\crow
\cmk{4}\enspace\ctk{+2.0k} \cq{\hlg{All good.} \hlp{Now, write the matrix in Python}} \textcolor{gray}{(length limit)}
& \cmk{4}\enspace\cnote{Student and teacher: 0/4.} Re-checking exhausts the remaining budget before the answer is emitted.
\cend
\cseg{Teacher continuation \hfill enlarged budget 8,192}{figteal}
\cmk{5}\enspace\ctk{+2.0k} \texttt{\cfence python} \ctk{+2.3k} \texttt{</think>}
& \cmk{5}\enspace\cnote{Teacher: 4/4.} With more continuation budget, the teacher finishes in 2.4k--3.8k tokens.
\cend
\end{casetable}
\end{casebox}
\casecaption{Late intervention on the same Survo puzzle. With $\ell_u=2048$, the teacher and student solve the remaining cells from the hand-off state but exhaust the original budget before emitting the answer. Extending the teacher's continuation budget restores completion.}{fig:case_late}

\subsection*{Remaining budget and response completion}
\begin{tcolorbox}[enhanced,colback=softgray,colframe=pivotRule,boxrule=.5pt,
  arc=1.2mm,left=3mm,right=3mm,top=2mm,bottom=2mm,
  fontupper=\fontsize{10}{12.5}\selectfont]
\setlength{\tabcolsep}{5pt}\renewcommand{\arraystretch}{1.2}
\begin{tabularx}{\linewidth}{@{}>{\raggedright\arraybackslash}Xcc@{}}
\textbf{Continuation from the late state} & \textbf{Budget} & \textbf{Completions}\\
\midrule
Student & 2,048 tokens & 0/4\\
Teacher & 2,048 tokens & 0/4\\
Teacher, enlarged budget & 8,192 tokens & \textcolor{figteal}{\textbf{4/4}}\\
\end{tabularx}
\end{tcolorbox}

\textbf{Observation.} The student resolves the remaining cells within 0.7k tokens, yet reaches the response limit before producing the requested answer. The teacher also fails under the 2,048-token continuation budget and succeeds in all four continuations when given 8,192 tokens.

\textbf{Interpretation.} The failure in this late-intervention example is consistent with insufficient budget for completing the response after solving the puzzle. A useful hand-off must leave time for verification and answer emission as well as for the remaining reasoning.

\end{document}